\documentclass[11pt,draftcls,onecolumn] {IEEEtran}
\ifCLASSINFOpdf
\else
\fi
\usepackage{amssymb,amsmath}
\usepackage[normalem]{ulem}
\usepackage[color,final]{showkeys}
\usepackage{multirow}
\usepackage{graphics}
\usepackage{graphicx}
\usepackage{subfigure}
\usepackage{epsfig}

\newcommand{\R}{{\mathbb{R}}}
\begin{document}

%
\title{FTVd is beyond Fast Total Variation regularized Deconvolution}
%
%
%

\author{Yilun~Wang,~\IEEEmembership{Member,~IEEE}
\thanks{Y. Wang is with the School of Mathematical Sciences, University of Electronic Science and Technology of China, Sichuan, 611731 China e-mail:yilun.wang@uestc.edu.cn}
}

%
%

\markboth{Journal of \LaTeX\ Class Files,~Vol.~6, No.~1, January~2007}%
{Shell \MakeLowercase{\textit{et al.}}: Bare Demo of IEEEtran.cls for Journals}
%



\maketitle

\begin{abstract}

In this paper, we revisit the ``FTVd" algorithm for Fast Total Variation Regularized Deconvolution, which has been widely used in the past few years.  Both its  original version implemented in the MATLAB software FTVd 3.0 and its related variant implemented in the latter version FTVd 4.0 are considered \cite{Wang08FTVdsoftware}.  We propose that the intermediate results during the iterations are the solutions of a series of combined  Tikhonov and total variation
regularized image deconvolution models and therefore some of them often have even better image quality than the final solution, which is corresponding to the pure 
total variation regularized model. 

\end{abstract}


%
\IEEEpeerreviewmaketitle

\section{Introduction}
%
%
%
%
Total variation regularized least-squares deconvolution is one of the most standard image processing problems, and the ``FTVd" package \cite{Wang08FTVdsoftware,Wang08FTVd} is among the most popular MATLAB softwares due to its outstanding computational efficiency.

We first briefly review the FTVd  as follows.
For simplicity,
the underlying images are assumed to have square domains, though all discussions can be
equally applied to rectangle domains.  Let $u^0\in\mathbb{R}^{n^2}$ be
an original $n\times n$ gray-scale image, $K\in\mathbb{R}^{n^2\times
n^2}$ represent a blurring (or convolution) operator,
$\omega\in\mathbb{R}^{n^2}$ be additive noise, and
$f\in\mathbb{R}^{n^2}$ an observation which satisfies the
relationship:
\begin{eqnarray}\label{Eq:Basic}
f=Ku^0+\omega.
\end{eqnarray}
Given $f$ and $K$, the image $u^0$ is recovered from the
model
\begin{eqnarray}\label{TV-reg}
\min_u \sum_{i=1}^{n^2}\|D_iu\|+\frac{\mu}{2}\|Ku-f\|_2^2,
\end{eqnarray}
where $D_iu \in \R^2$ denotes the
discrete gradient of $u$ at pixel $i$, and the sum $\sum\|D_iu\|$
is the discrete total variation (TV) of $u$.  Model (\ref{TV-reg}) is also widely
referred to as the TV/$L^2$ model. Here the TV is {\em isotropic} if the norm $\|\cdot\|$
in the summation is $2$-norm, and {\em anisotropic} if it is $1$-norm.  While FTVd 
applies to both isotropic and anisotropic TV,  we will only review the isotropic case,
$\|\cdot\|=\|\cdot\|_2$ because the treatment for the
anisotropic case is completely analogous.

As we have already known, FTVd is initially derived from the well-known operator-splitting and quadratic
penalty techniques in optimization as follows. At each pixel an
auxiliary variable $\mathbf{w}_i\in\mathbb{R}^2$ is introduced to
transfer $D_iu$ out of the non-differentiable term $\|\cdot\|_2$ as follows.
\begin{eqnarray}\label{Prob:constraint}
\min_{\mathbf{w}, u
}\sum_i\|\mathbf{w}_i\|_2
+ \frac{\mu}{2}\|Ku-f\|_2^2, \quad s.t. \quad  \mathbf{w}_i=D_iu
\end{eqnarray}
Notice that the main purpose of the introduction of $\mathbf{w}$ is for the operator splitting.
Then FTVd aims to get an unconstrained version of \eqref{Prob:constraint} for easiness of computation and adopts different methods to deal with the constraint $ \mathbf{w}_i=D_iu$ in its different versions. In the FTVd 3.0 version, the simple quadratic penalty scheme was adopted, 
yielding the following approximation model of \eqref{Prob:constraint}:
\begin{eqnarray}\label{AppProb}
\min_{\mathbf{w}, u
} \mathcal{Q}_{\mathcal{A}}(u,\mathbf{w},\beta)
\end{eqnarray}
with a sufficiently large penalty parameter $\beta$, where \begin{eqnarray*}
\mathcal{Q}_{\mathcal{A}}(u,\mathbf{w},\beta)&\doteq& \sum_i\|\mathbf{w}_i\|_2
+ \frac{\beta}{2}\sum_i\|\mathbf{w}_i-D_iu\|_2^2\\
&+& \frac{\mu}{2}\|Ku-f\|_2^2.
\end{eqnarray*}  This type of
quadratic penalty approach can be traced back to as early as Courant
\cite{Courant1943} in 1943. The motivation for this formulation is that when
either one of the two variables $u$ and $\mathbf{w}$ is fixed,
minimizing the function with respect to the other has a closed-form
formula with low computational complexity and high numerical
stability for a given $\beta$, resulting in the following  alternative minimization scheme.
\begin{eqnarray} \label{eq:AM}
\left \{ \begin{array}{c}
         \mathbf{w}^{k+1}\leftarrow \arg \min_{\mathbf{w}}\mathcal{Q}_{\mathcal{A}}(u^{k},\mathbf{w},\beta) \\
        u^{k+1}\leftarrow \arg \min_{u}\mathcal{Q}_{\mathcal{A}}(u,\mathbf{w}^{k+1},\beta) \\  \end{array} \right.
\end{eqnarray}
 The detailed formulations of the close form solution of each  subproblem can be referred to \cite{Wang08FTVd}. For a given $\beta$, this alternative minimization will converge to the solution of (\ref{AppProb}).

In order to make the $u$-solution of (\ref{AppProb}) be or very close to the solution of the original model \eqref{TV-reg},
 $\beta$ needs to  be a very large positive number.  It has been showed  that, for any fixed $\beta$, the algorithm of solving \eqref{AppProb} via
minimizing $u$ and $\mathbf{w}$ alternately has faster convergence when
$\beta$ is small, but slower when $\beta$ is large. In order to faster get the solution of (\ref{AppProb}) for large $\beta$, FTVd (implemented in FTVd 3.0 version)
adopted  the continuation scheme, i.e. the $\beta$ increases gradually from a small number to a larger one, where  earlier subproblems \eqref{eq:AM} corresponding to smaller $\beta$ values can be solved
quickly, and the solution is used as the initial guess for the later subproblems \eqref{eq:AM}. 


As an alternative for the continuation scheme, 
the newer version of FTVd (FTVd 4.0 version) augmented  the quadratic penalty term  by adding a lagrangian term, i.e. making use of the following augmented lagrangian function. 
%
\begin{eqnarray}\label{eq:augLag}
\mathcal{L}_{\mathcal{A}}(u,\mathbf{w},\lambda)&\doteq& \sum_{i}(\|\mathbf{w}_i\|-\lambda_i^{\mathrm{T}}(\mathbf{w}_i-D_i x)\\\nonumber &+&\frac{\beta}{2}\|\mathbf{w}_i-D_i x\|^2)+\frac{\mu}{2}\|Ku-f\|^2
\end{eqnarray}
We have the following iterative framework to solve it.
\begin{eqnarray} \label{eq:ADM1}
\left \{ \begin{array}{c}
         (\mathbf{w}^{k+1}, u^{k+1}) \leftarrow \arg \min_{\mathbf{w}, u}\mathcal{L}_{\mathcal{A}}(\mathbf{u},\mathbf{w},\lambda^{k}) \\
         \lambda^{k+1} \leftarrow \lambda^{k}-\beta(\mathbf{w}^{k+1}-Du^{k+1}) \\
        \end{array} \right.
\end{eqnarray}
It is well-known that the presence and iterative updates of multiplier $\lambda$ avoids $\beta$ explicitly going to infinity and guarantees convergence of \eqref{eq:ADM1} to a solution
of the original TV model (\ref{Prob:constraint}). 
%
For the sake of computational easiness,
\eqref{eq:ADM1} can be further written as the alternative direction method (ADM, in short), which
 was studied extensively in optimization and variational analysis, following the pioneer work \cite{Galowinski75ADM,Gabay76dual,Glowinski89ADM,Douglas56splitting}.
 ADM  applied to (\ref{eq:augLag}) or \eqref{eq:ADM1} yields the following iterative scheme:

\begin{eqnarray} \label{eq:ADM}
\left \{ \begin{array}{c}
         \mathbf{w}^{k+1}\leftarrow \arg \min_{\mathbf{w}}\mathcal{L}_{\mathcal{A}}(u^{k},\mathbf{w},\lambda^{k}) \\
        u^{k+1}\leftarrow \arg \min_{u}\mathcal{L}_{\mathcal{A}}(u,\mathbf{w}^{k+1},\lambda^{k}) \\
        \lambda^{k+1} \leftarrow \lambda^{k}-\beta(\mathbf{w}^{k+1}-Du^{k+1})
        \end{array} \right.
\end{eqnarray}
Here the resulting subproblems do  have closed-form solutions in our context of image deconvolution. In general, FTVd 4.0 has a faster convergence than FTVd 3.0.
In \cite{Goldstein09splitBregman},
the authors derived (\ref{eq:ADM}) from the Bregman iterative method \cite{Osher05Bregman}, and the equivalence between split bregman and with the ADM 
under certain conditions is established and analyzed in \cite{Esser09Equivalence,Wu10ALM}.
\subsection{Contributions}
The main viewpoint of the paper is that
solving  the problem \eqref{TV-reg}  in FTVd is through solving a series of  the combined Tikhonov and total variation regularized image deconvolution models, and the details will be given in Section \ref{sec:contribution}.
\begin{eqnarray}\label{eq:combine}
\min_{u=u_1+ u_2
} \texttt{TV}(u_1) +\frac{\beta}{2} \texttt{Tikhonov}(u_2)
+ \frac{\mu}{2}\|Ku-f\|_2^2,
\end{eqnarray}
Therefore, the intermediate results of FTVd are corresponding to different size of $\beta$ and the final solution of FTVd is corresponding to a huge $\beta$ where the TV regularization dominates and the Tikhonov regularization is almost ignorable. 
%
Thus, some of the intermediate solutions of FTVd 
might be of higher recovered image quality such as reduced staircase artifacts than its final solution, which is also demonstrated by numerical experiments. 
%


\subsection{Paper Organization}
 The organization of the paper is as follows. We will further explain  why FTVd is beyond the signal total variation regularized model 
 in Section \ref{sec:contribution}.
In Section \ref{sec:numExp}, several numerical experiments will verify our point of view. In the end, the conclusion and future work will be given.

\section{FTV\lowercase{d} involves both Total Variation Regularization and Tikhonov Regularization} \label{sec:contribution}
FTVd is based on the  constrained version of the original TV model  \eqref{Prob:constraint} resulted from  operator splitting. 
In order to get an equivalent unconstrained version of \eqref{Prob:constraint},  the quadratic penalty method is adopted in FTVd 3.0 together with continuation scheme,  and then augmented with a lagrangian term in FTVd 4.0, to avoid explicitly performing the continuation scheme. While most of existing work consider the computational efficiency of FTVd for the original TV model \ref{TV-reg},  we will consider FTVd from image decomposition point of view to examine the recovery image quality during the iterations proceed in this paper.  

Let $u=u_1+u_2$ and $D_i u_1 =\mathbf{w}$,  then $D_i u_2=D_i u-\mathbf{w}_i$, where $\mathbf{w}$ is the auxiliary variable.  Equation (\ref{AppProb}) based on
operator splitting and the quadratic penalty adopted in FTVd 3.0, can be rewritten as

\begin{eqnarray}\label{AppProb2}
\min_{u_1, u_2
}\sum_i\|D_iu_1\|_2
&+& \frac{\beta}{2}\sum_i\|D_iu_2\|_2^2 \\\nonumber
&+& \frac{\mu}{2}\|K(u_1+u_2)-f\|_2^2,
\end{eqnarray}
Here 
we consider the recovered image $u$ as a sum of
two components: a piecewise constant component $u_1$ and a smooth component $u_2$.
Correspondingly,  \eqref{AppProb}  is considered as a combination of Tikhonov regularization and total variation regularization. Therefore, the surrogate model \eqref{AppProb} is expected to be more flexible than the original model \eqref{TV-reg}, and its solution expect to have smaller staircase artifacts, which are inherent in the single total variation regularization model \cite{Gholami2013TVTik}.
Specifically, Tikhonov regularization preserves small derivative coefficients  while severally penalizing the spikes, hence recovering smooth regions while smoothing discontinuities in the regularized solution. In contrast, total variation regularization preserves the spikes while severely penalizing the small coefficients, likely changing the smooth regions in final solution into staircases.

There have existed several similar models \cite{Chambolle97Highorder} based on image decomposition. In order to better preserve corners and edges, or reduce staircasing of the total variation regularization, an infimal-convolution functional
\begin{eqnarray*}
\min_{u_1+u_2=u
}\int_{\Omega} |\nabla u_1|+ \alpha|\nabla(\nabla u_2)| dx +\frac{\mu}{2}\|Ku-f\|_2^2,
\end{eqnarray*}
was proposed in \cite{Chambolle97Highorder} and proved to be practically effective for images containing various grey levels as well as edges and corners. A modified form was proposed in \cite{Chan10FourthTV}, where the regularization term is of the form
$$\min_{u_1+u_2=u}\int_{\Omega} |\nabla u_1|+\alpha |\triangle u_2|dx +\frac{\mu}{2}\|Ku-f\|_2^2,;$$
i.e., the second derivative is replaced by the Laplacian, and a dual method for its numerical realization is derived.  This above idea could certainly be used in many other settings, when a signal or an
image presents various types of characteristic features that need to be preserved
and reconstructed.


  The novelty of this paper is that while the quadratic penalty and the continuation scheme are originally proposed mainly for computational efficiency,  we reconsider them from the viewpoint of image decomposition, 
  as  showed in (\ref{AppProb2}), which is equivalent with (\ref{AppProb}). The quadratic penalty corresponding to the Tikhonov regularization of $u_2$.  
%
   In FTVd 3.0, 
the intermediate results corresponding different $\beta$ is the solution of \eqref{AppProb} or \eqref{AppProb2}, which is regularized by both total variation and Tikhnov regularization. Only when $\beta$ is large enough, its solution is well approximating the true solution of the original TV model (\ref{TV-reg}), because  the Tiknonov regularization is approaching to null.   Thus, the continuation procedure is in fact also a procedure of testing different $\beta$ values to find out the one corresponding to the best recovery quality. 
In particular,  we record all the 
  intermediate results during the continuation procedure, pick the one of the best visual quality.  We can even stop increasing $\beta$ if we think the recovery quality is no longer improving.  




FTVd 4.0  augmented the quadratic penalty by adding a lagrangian term, i.e. making use of the augmented lagrangian method (ALM) \cite{Nocedal_book}. 
The explicitly increasing $\beta$ in FTVd 3.0 is replaced by estimating the multiplier for a fixed $\beta$, as did in  \eqref{eq:ADM1}, which can also be considered as a way of implicitly increasing $\beta$ in context of quadratic penalty.  
 Due to the implicit way of increasing $\beta$ in context of quadratic penalty, 
%
%
FTVd 4.0 also enjoys the advantages of a combination of Tikhonov and total variation regularizations for reconstruction of piecewise-smooth signals during its intermediate iterations. In particularly, in terms of the ALM, the intermediate solutions correspond to the mixed model \eqref{AppProb2} of a sequence of $\beta$ gradually increased from small to large. So we expect some $u$ results of certain intermediate iterations of FTVd 4.0 to have better  recovery quality than the final solution, due to their  corresponding to an appropriate balance 
between total variation and Tikhonov regularization. Here by ``the better image quality", we mainly mean reduced stair-case effects, which are inherent in the solution of the pure total variation model. 
%

%
%

For further computational convenience,  the alternating direction method (ADM) is then applied to ALM \eqref{eq:ADM1}, resulting to  alternatively minimizing $u$ and $\mathbf{w}$, as showed in \eqref{eq:ADM}. In such cases, we still kept all the intermediate $u$ solution and compared them with the final solution. The better image quality of intermediate results in terms of reduced staircase effects are demonstrated in the following numerical experiments.

\section{Numerical Experiments} \label{sec:numExp}
In this section, we present detailed numerical results to show the intermediate results and final solutions of FTVd 3.0 and
FTVd 4.0, respectively. The intermediate results can be obtained and kept without extra efforts as FTVd is performed. The main purpose is to show that some intermediate results empirically indeed have better
recovery quality than the final solution, at least in terms of slightly higher SNR and visually significant reduced staircase artifacts. The better quality can be also quantified using  non-reference image quality analyzers  \cite{Mittal12QualityEvaluator,Mittal13ImageQualityAnalyzer}. In this paper, we only use the SNR and eyeball observation due to the conciseness of the paper.   


\subsection{Experimental Settings}
Due to the shortness of the paper, we only show our results on the test image ``Barbara" with size $512\times 512$,  which has a nice mixture of detail, flat regions, shading area, and texture and  is suited for our experiments. Its grey levels are  normalized to $[0, 1]$.

FTVd was implemented in MATLAB \cite{Wang08FTVdsoftware}. The blurring kernel is generated by the MATLAB function ``fspecial" and all blurring effects were then generated using the MATLAB
function ``imfilter" with periodic boundary conditions. In all our experiments, we take the ``average" convolution kernel of size $9 \times 9.$  The additive noise used was Gaussian noise with
zero mean and standard deviation $\sigma=10^{-2}$.

The parameter $\mu$ controls the amount of regularization
applied to the squared $\ell_2$-distance between $Ku$ and $f$.  The issue of how to optimally select $\mu$ is important \cite{Strong-Chan-03,Wen09TVParaSel,Jin2010HPR},
but is outside of the scope of this work. FTVd adopts the empirical formula $\mu = 0.05/\sigma^2$, where $\sigma$ is the standard deviation of the additive Gaussian noise.


For FTVd 3.0, the quadratic penalty parameter $\beta$ starts from initial small positive values to large ones, as the continuation scheme proceeds. Here FTVd has $11$ outer iterations corresponding to the $\beta$-sequence $ \{2^0, 2^1, . . . , 2^{10}\}$.
We preserve all the intermediate results corresponding to the continuation steps and the final solution. Here for each intermediate  $\beta$,  we use the same  tight stopping tolerance as the final one when solving \eqref{eq:AM}, because we also care about the quality of the intermediate solutions. In the original implementation of FTVd 3.0,  the intermediate $\beta$ used a looser stopping tolerance, because they only cared about the final solution and this way a faster convergence can be achieved.

In FTVd 4.0, the $\beta$ is fixed and the default value is $10$.  As \eqref{eq:ADM} showed, the continuation scheme is replaced by the method of multipliers. The updating of multiplier $\lambda$ plays the similar role as increasing $\beta$ in FTVd 3.0, but achieve a faster convergence. We also record all the intermediate results during the iterations.



\subsection{Numerical Results}

\begin{figure}[h]
    \centering
     \includegraphics[width=0.2\textwidth]{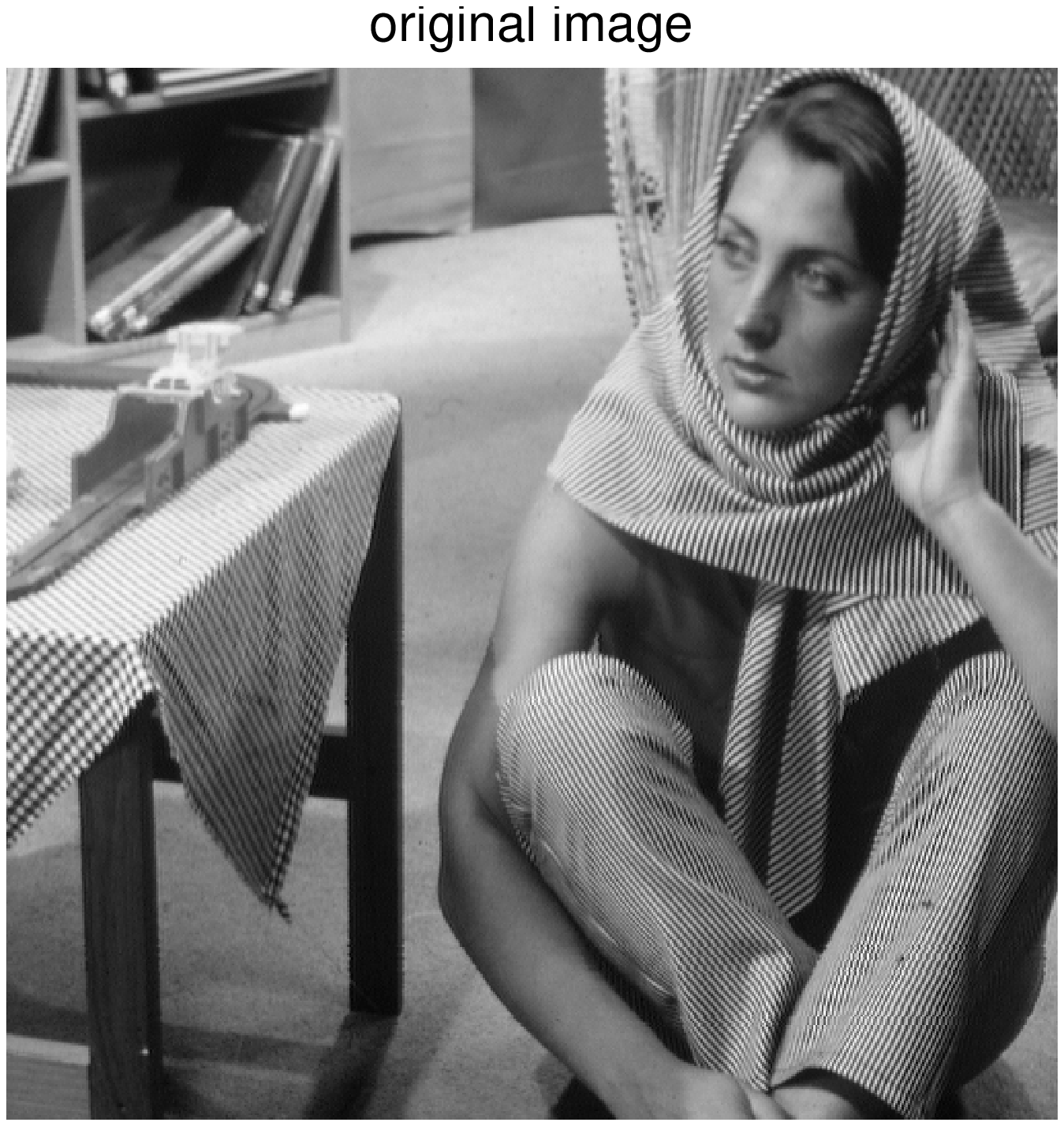}
       \includegraphics[width=0.2\textwidth]{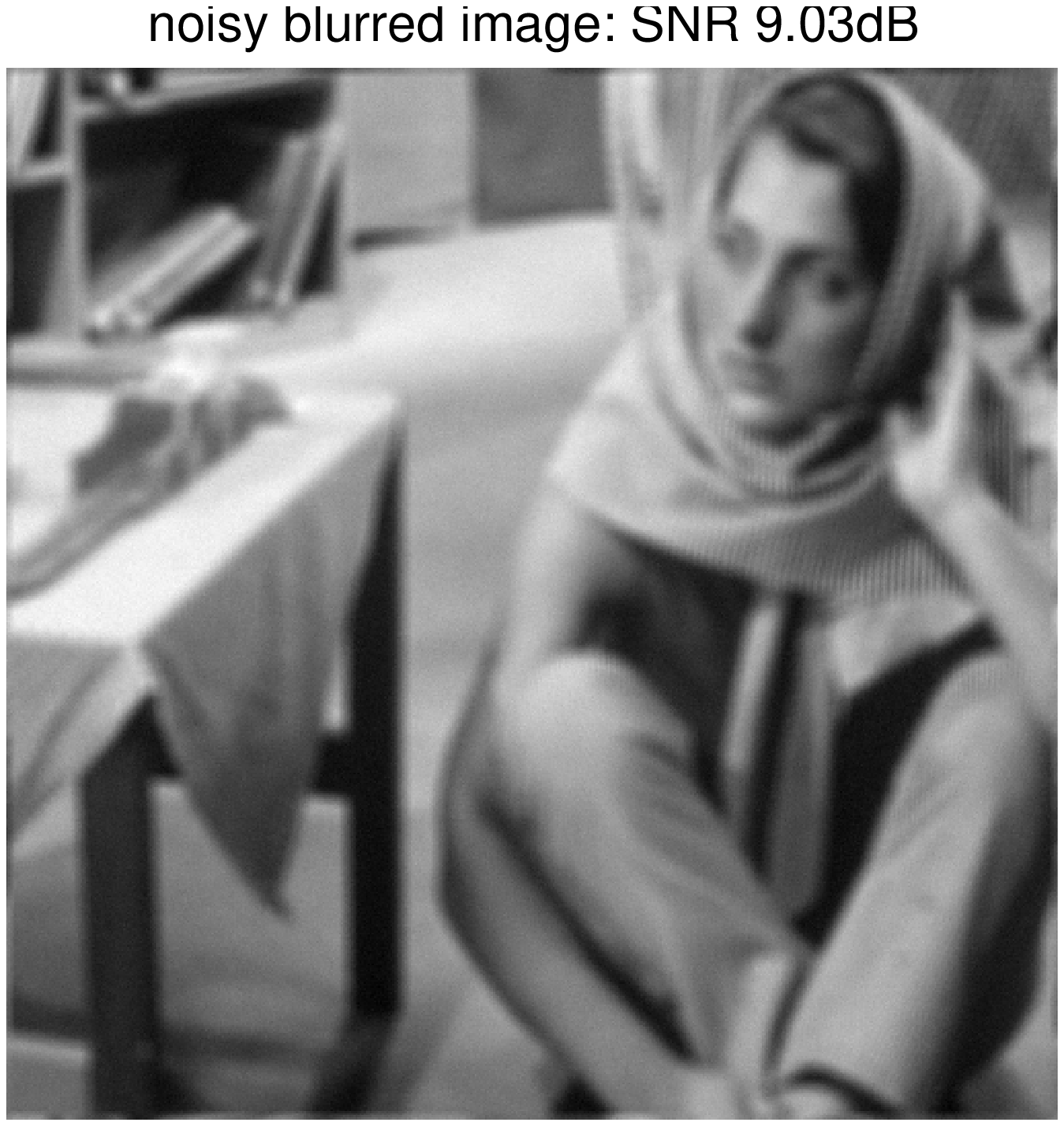}\\\vspace{-0.1cm}
       \includegraphics[width=0.2\textwidth]{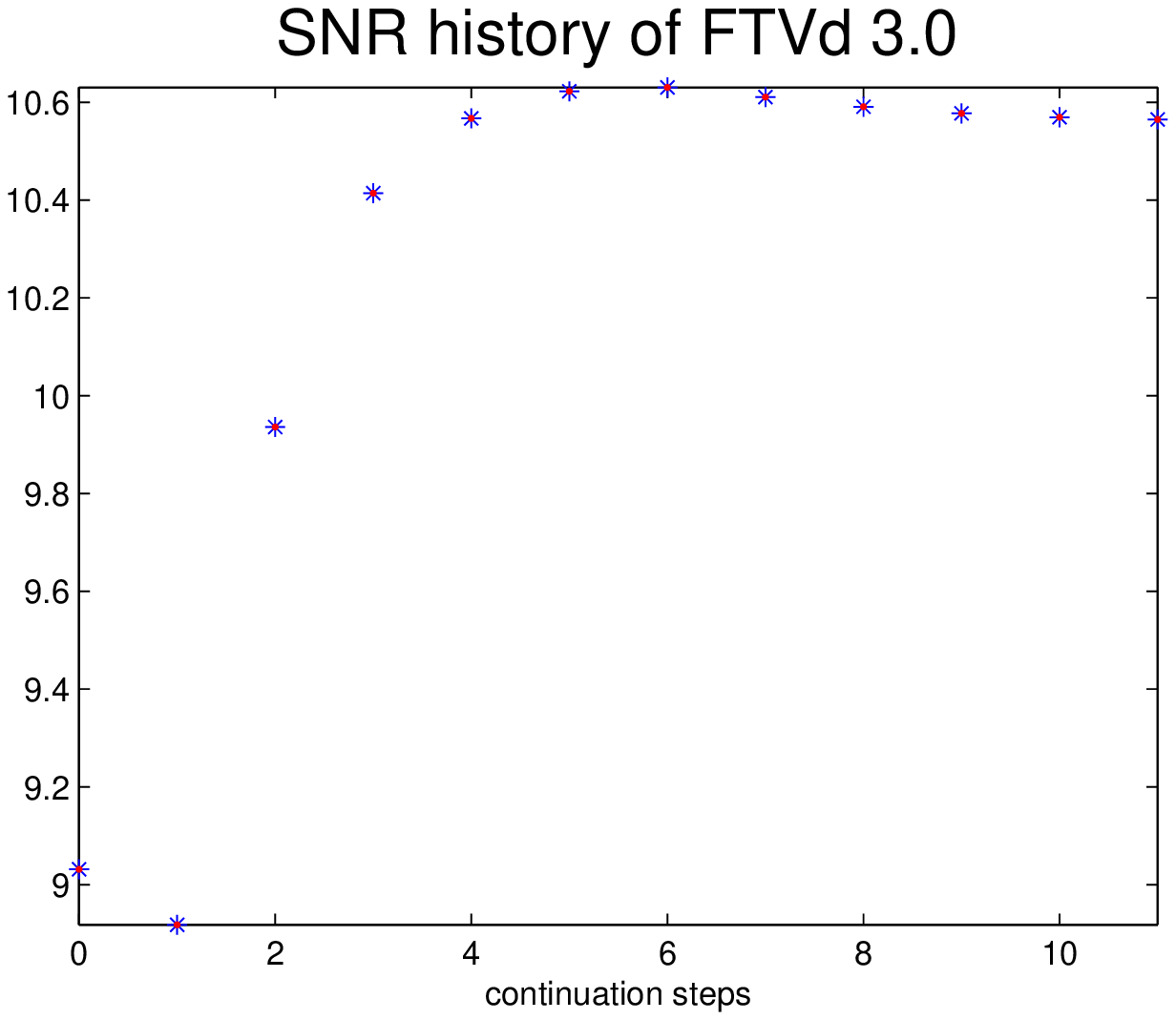}
      \includegraphics[width=0.2\textwidth]{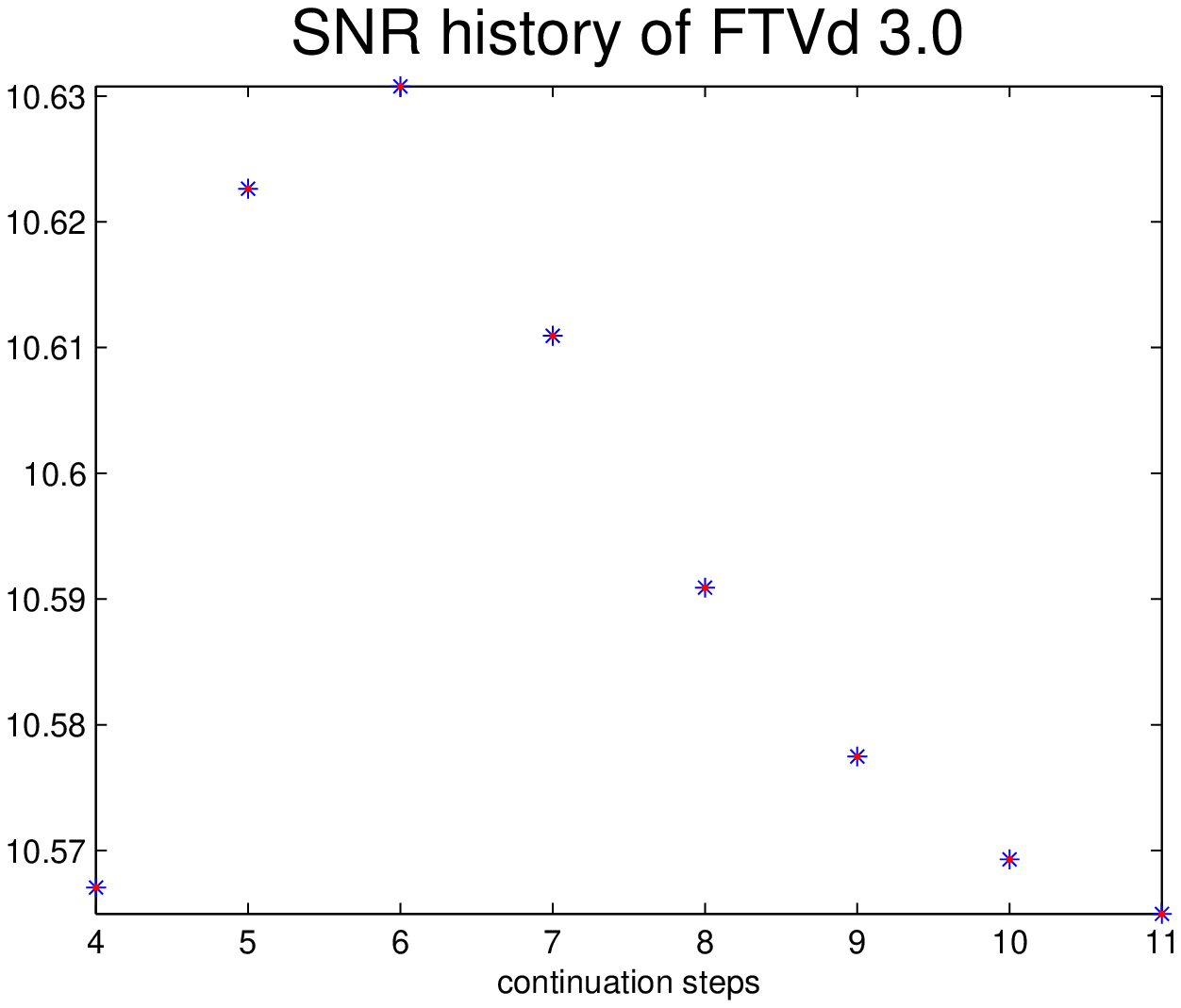}\\
     \includegraphics[width=0.2\textwidth]{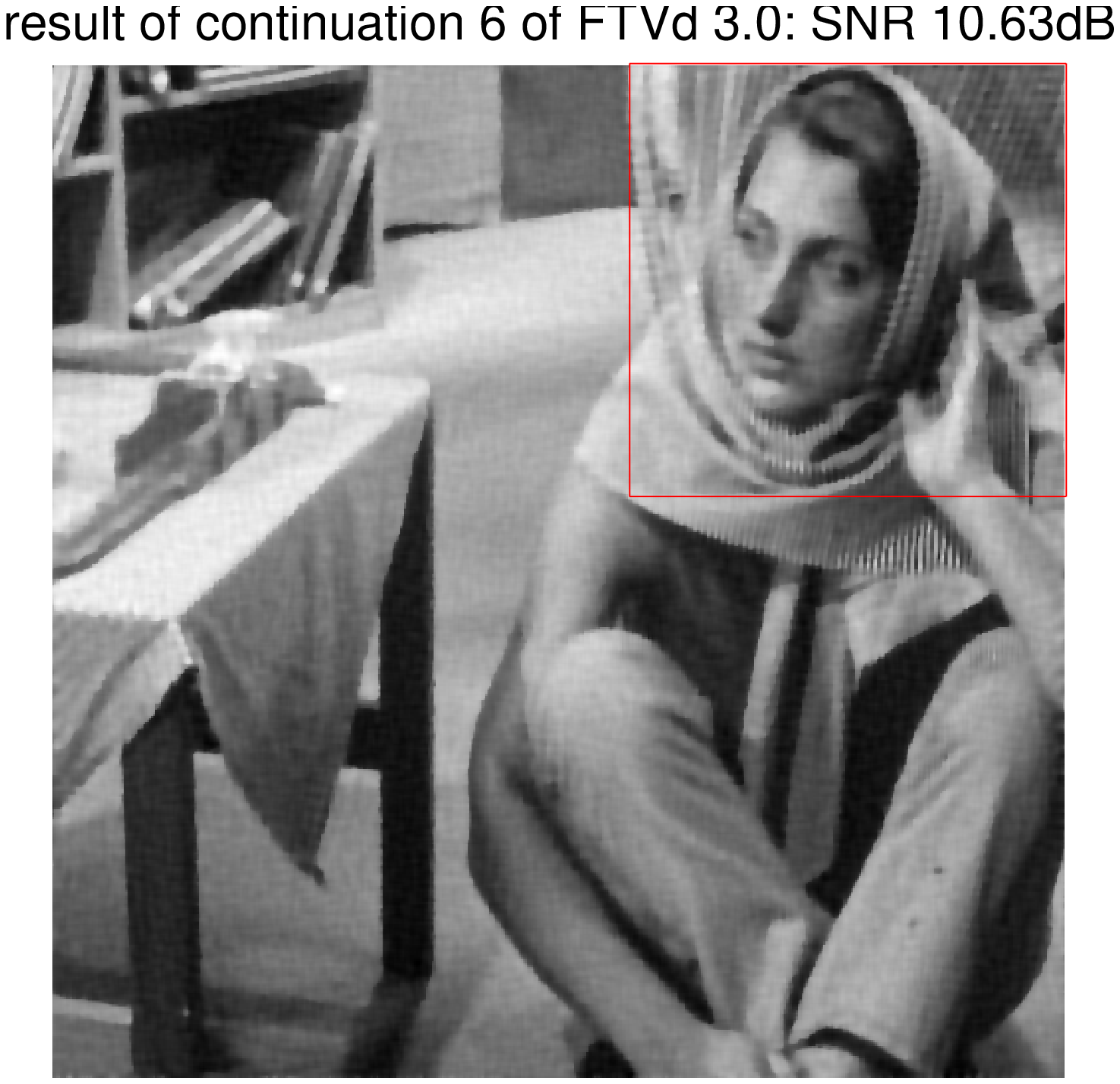}
    \includegraphics[width=0.2\textwidth]{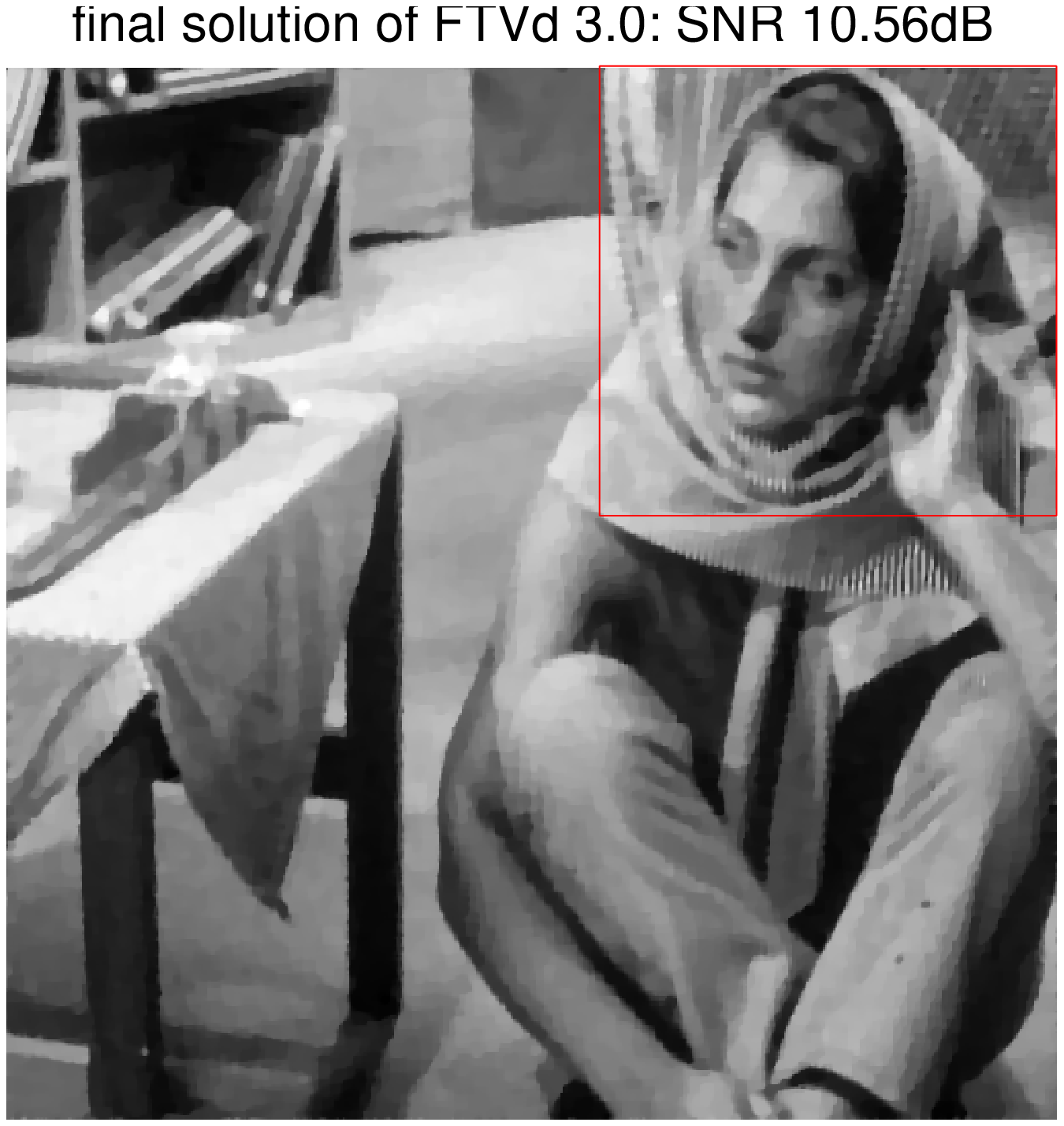}\\\vspace{-0.3cm}
     \includegraphics[width=0.25\textwidth]{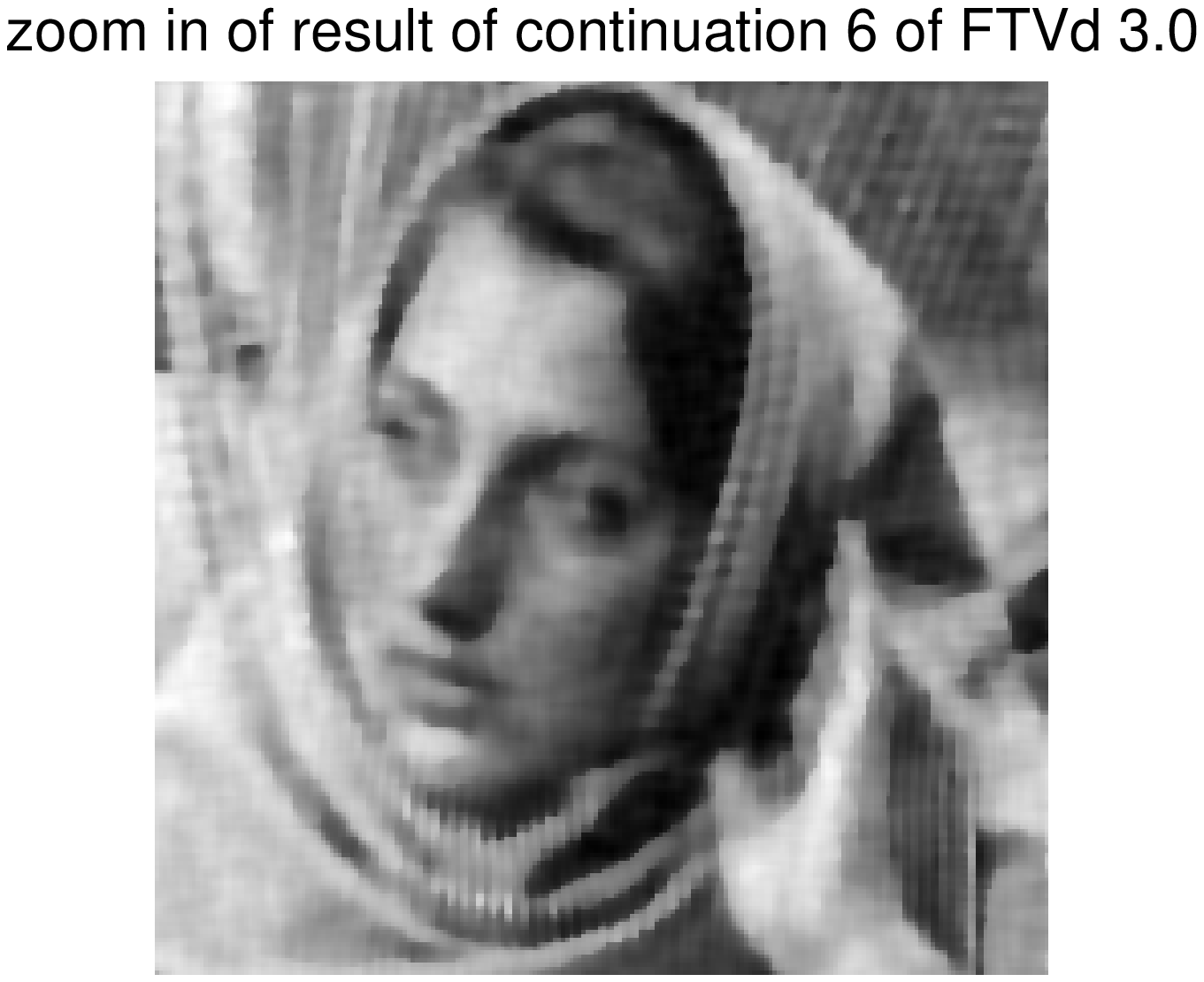}\hspace{-0.4cm}
    \includegraphics[width=0.25\textwidth]{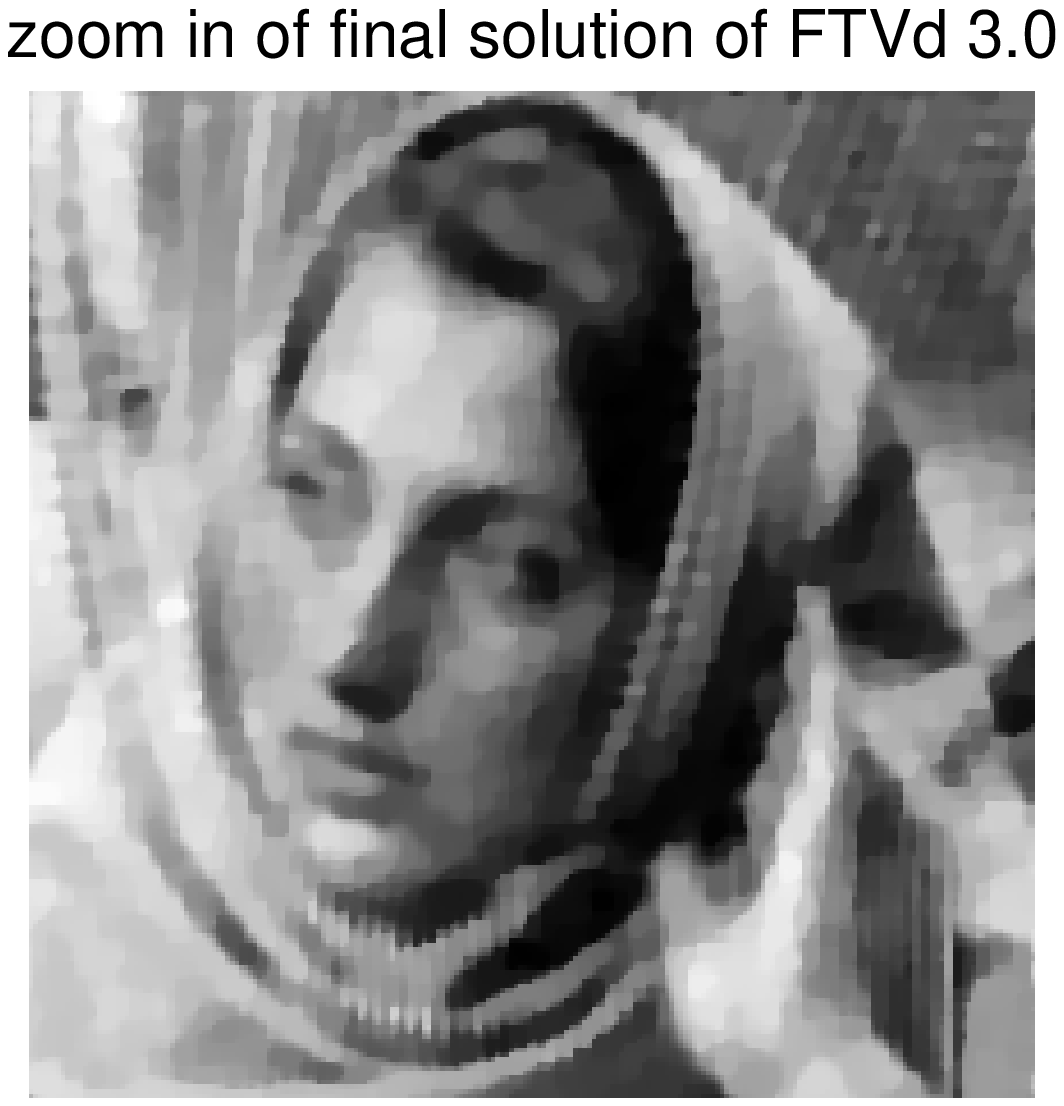}\\\vspace{-0.5cm}
    \caption{Image deblurring experiment of FTVd 3.0. Top left: original image. Top right: blurred noisy image. Second row: SNRs of the results of each continuation step. Left of third row: intermediate result of highest SNR. Right of third row: the final solution. Four row: zoom in displays. We can see the intermediate result of the 6-th continuation step has slightly high SNR and significant reduced staircase than the pure TV solution. }
    \label{fig:Barbara3}
\end{figure}

Figures \ref{fig:Barbara3} is the result of FTVd 3.0 run on the image
Barbara.
%
The left subfigure of the first row is the original image and the right one is the blurry and noisy one which is  $f$.
The SNRs of the initial guess ($f$, here) and the intermediate results of every continuation step including the last step are showed in the left subfigure of the second row.
The right subfigure is the truncated version for better display. We can see that SNRs increase dramatically first as the continuation proceeds and the total variation regularization begins to play a more and more important role,
then begin to slightly decrease in the last few continuation steps where the total variation dominates too much over Tikhonov regularization. While the decreasing of SNR in the latter part of the continuation procedure is not very significant,
we can see obvious differences visually as showed in the third and fourth rows. Specially, we plot the intermediate result of largest SNR (corresponding to the sixth continuation step) on the left subfigure of the third row, and the final solution (corresponding to the final continuation step) on the right subfigure. We can see that the intermediate result corresponding to the balanced combined TV and Tikhonov regularizations \eqref{AppProb2} has significant reduced staircase artifacts than the final solution, which is expected to be or very close to the solution of the pure TV regularization model \eqref{TV-reg}. The reduced staircase effect can be better observed in the zoom in display of the specified part, as showed in the fourth row. Reducing staircase artifacts is a very important task in computer vision \cite{Xu14TGV}.

\begin{figure}[h!]
    \centering
     \includegraphics[width=0.2\textwidth]{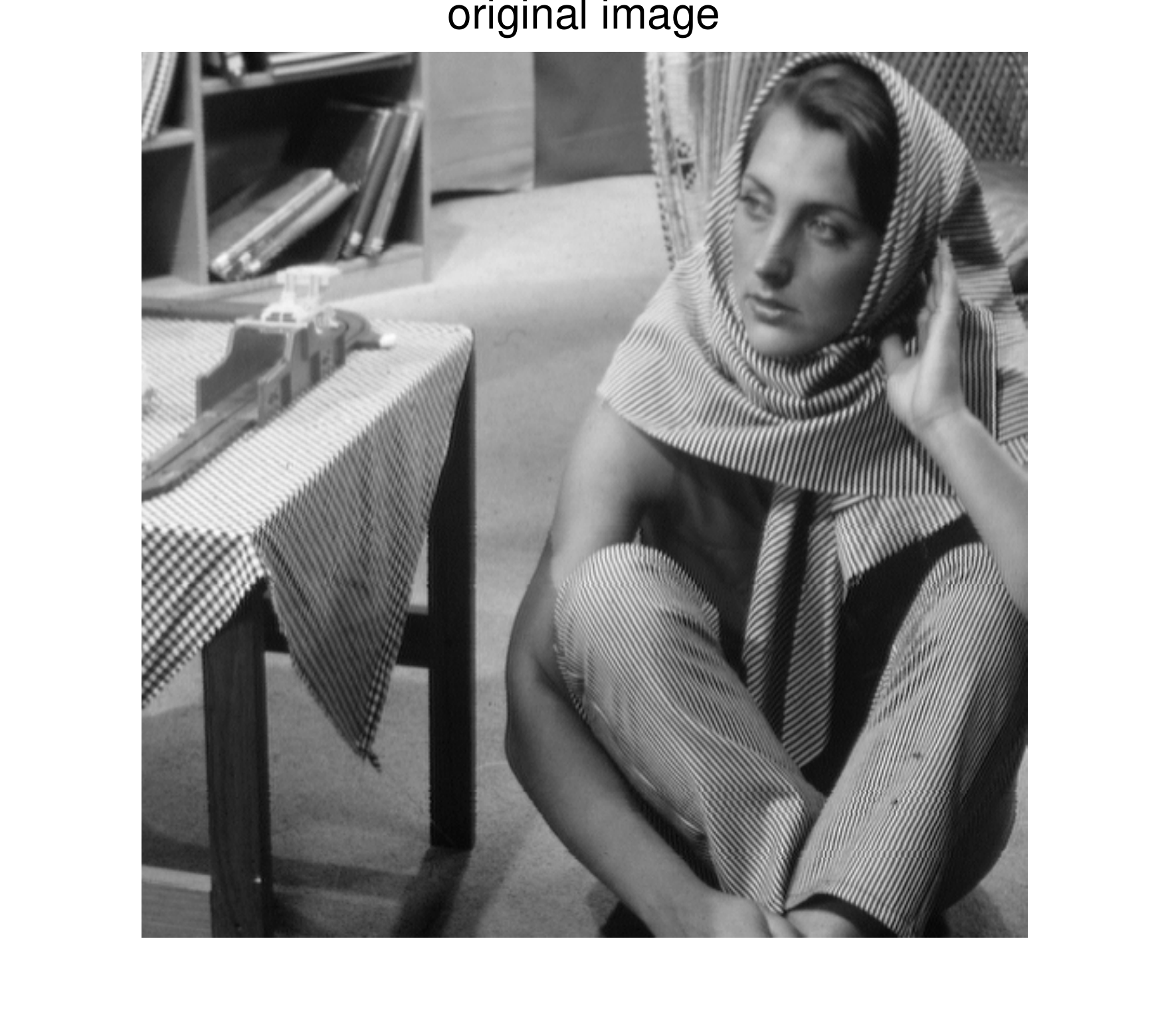}
       \includegraphics[width=0.2\textwidth]{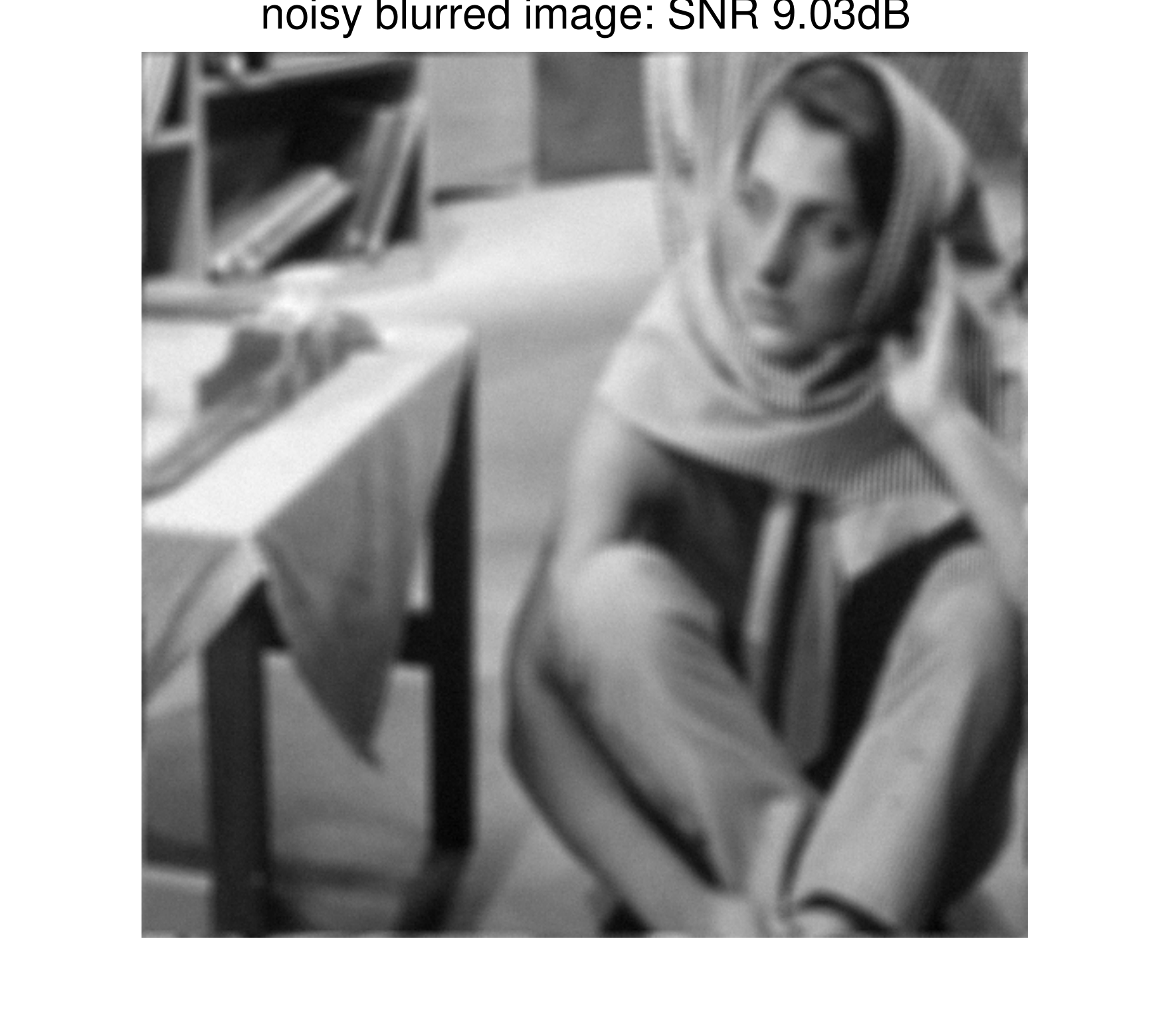}\\
        \includegraphics[width=0.2\textwidth]{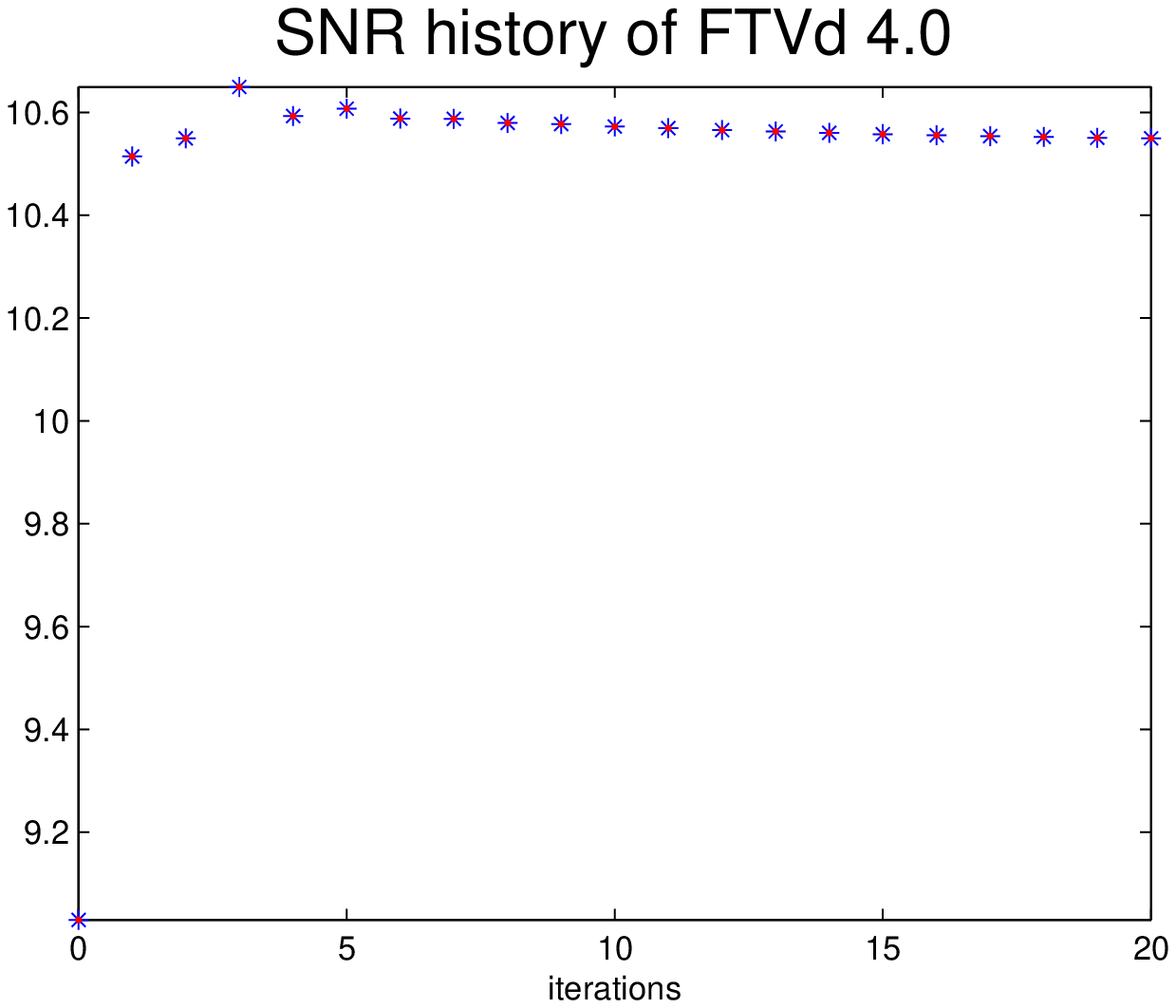}
      \includegraphics[width=0.2\textwidth]{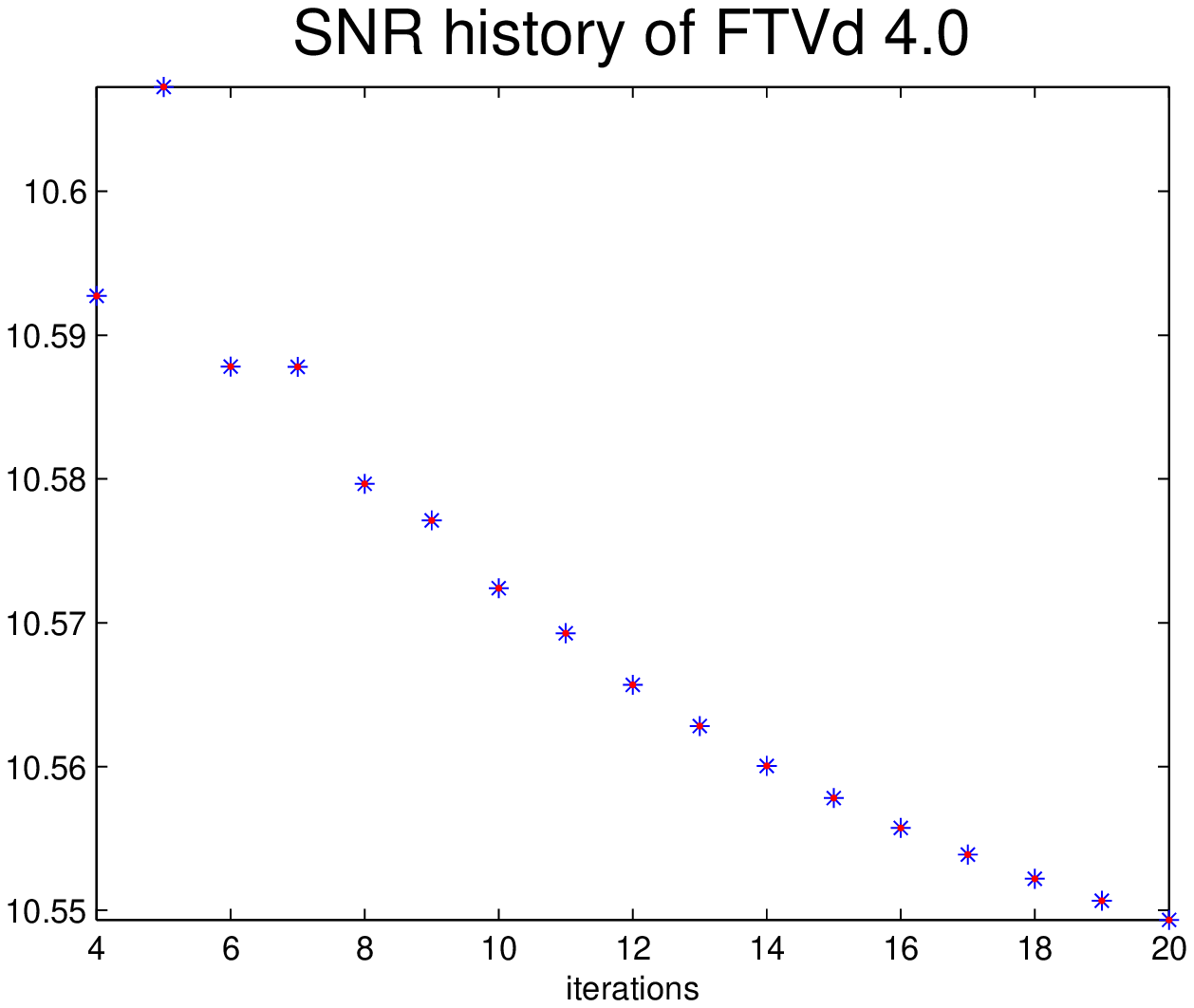}\\
     \includegraphics[width=0.2\textwidth]{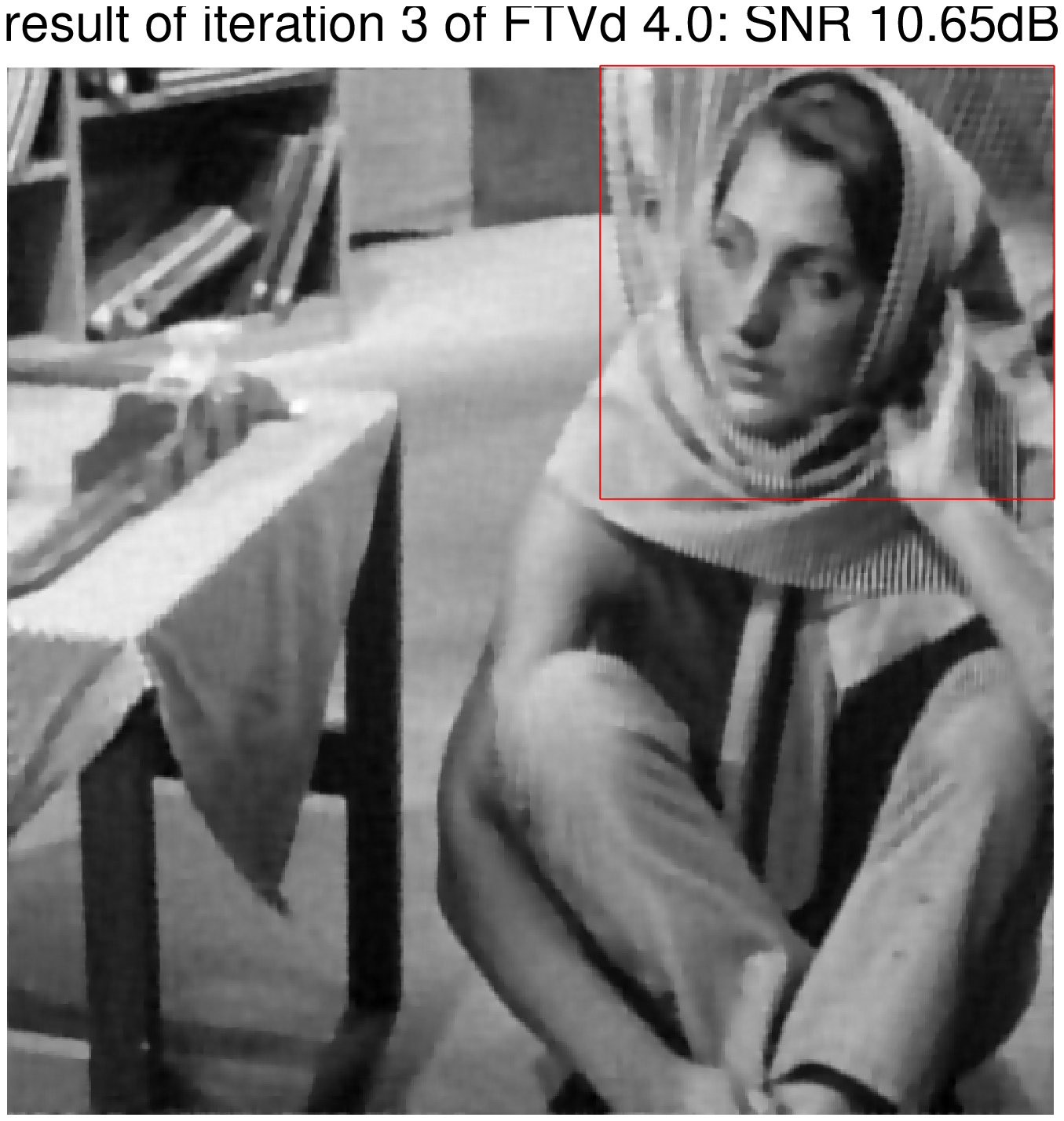}
    \includegraphics[width=0.2\textwidth]{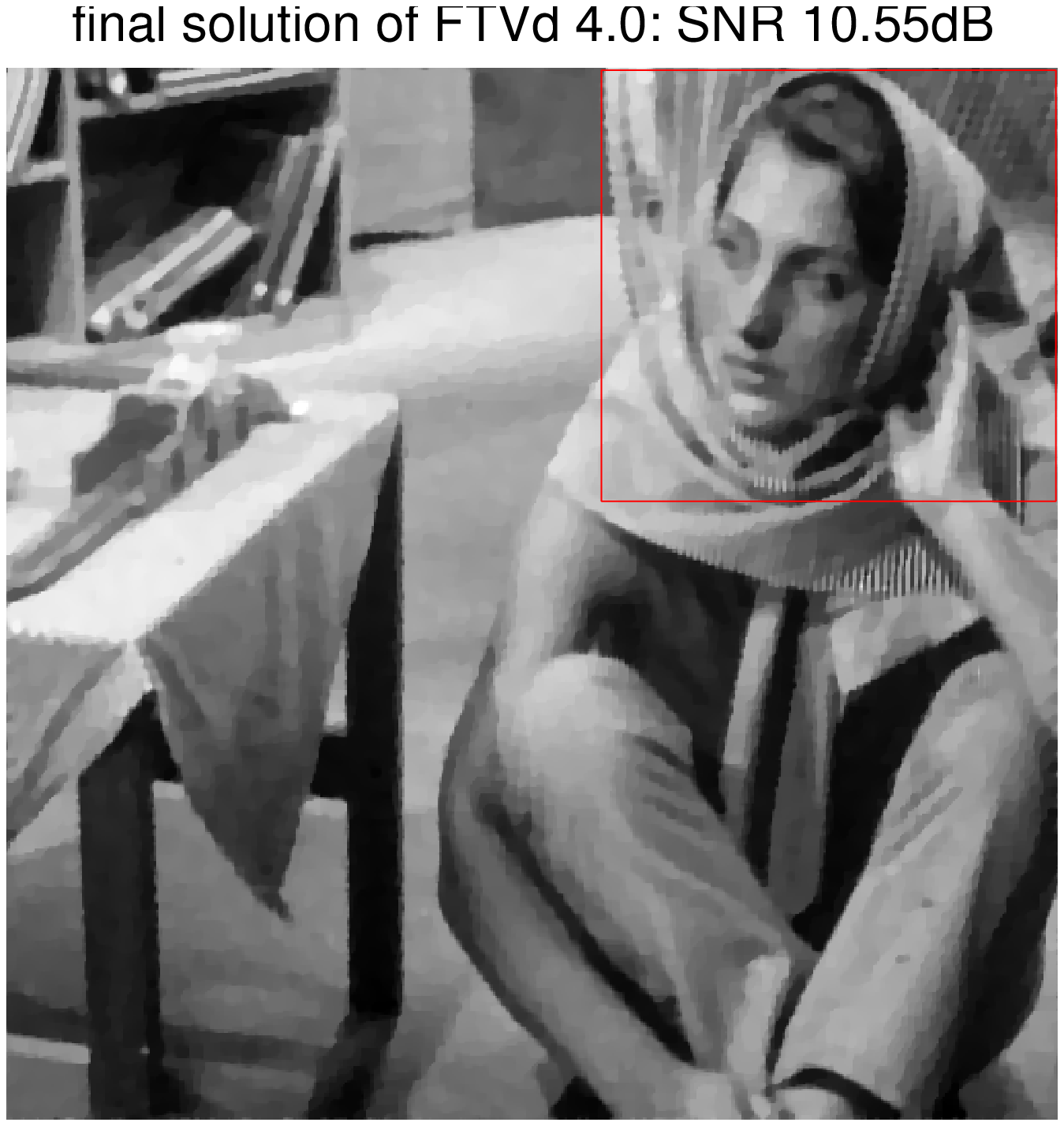}\\
     \includegraphics[width=0.25\textwidth]{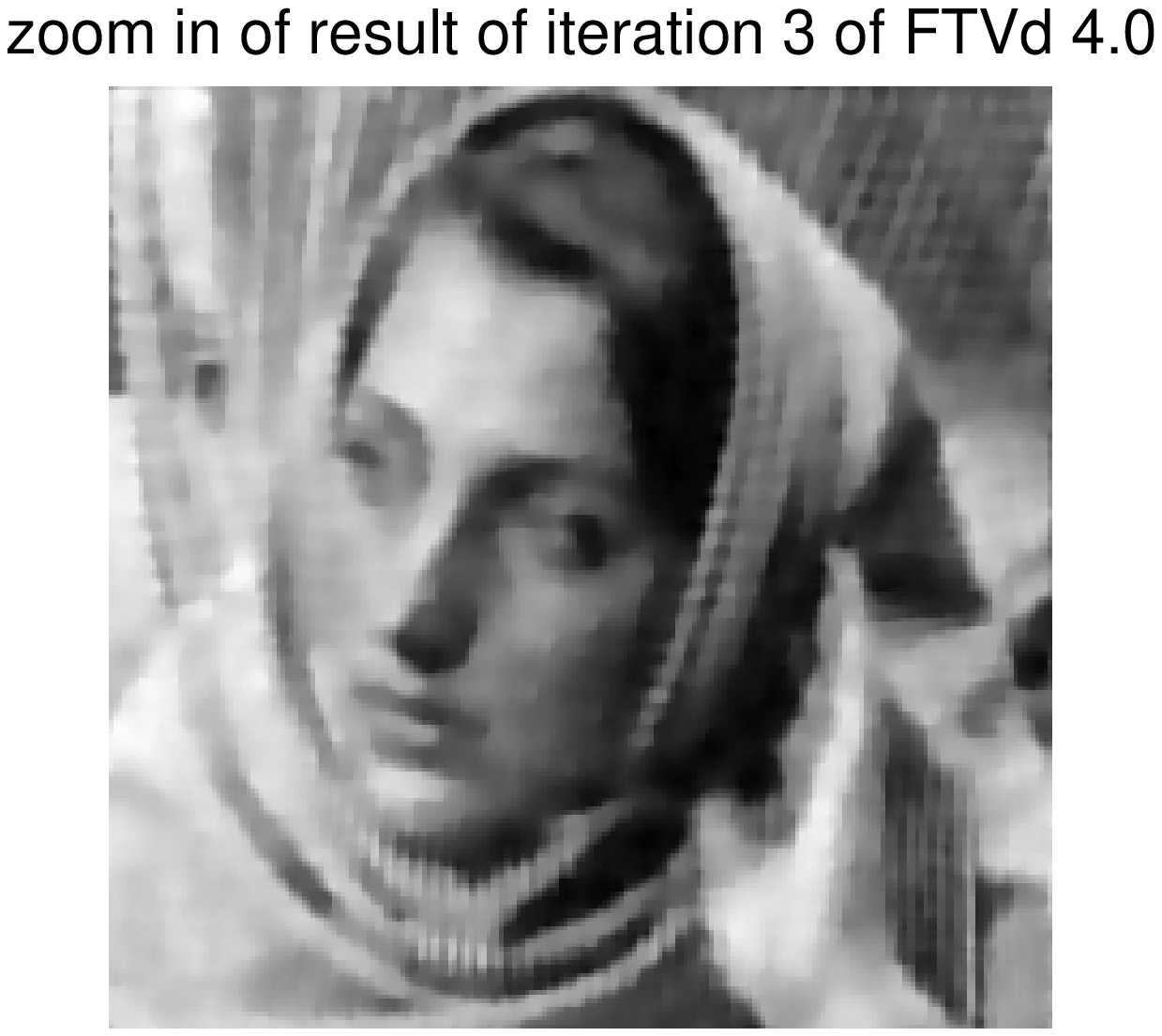}
    \includegraphics[width=0.25\textwidth]{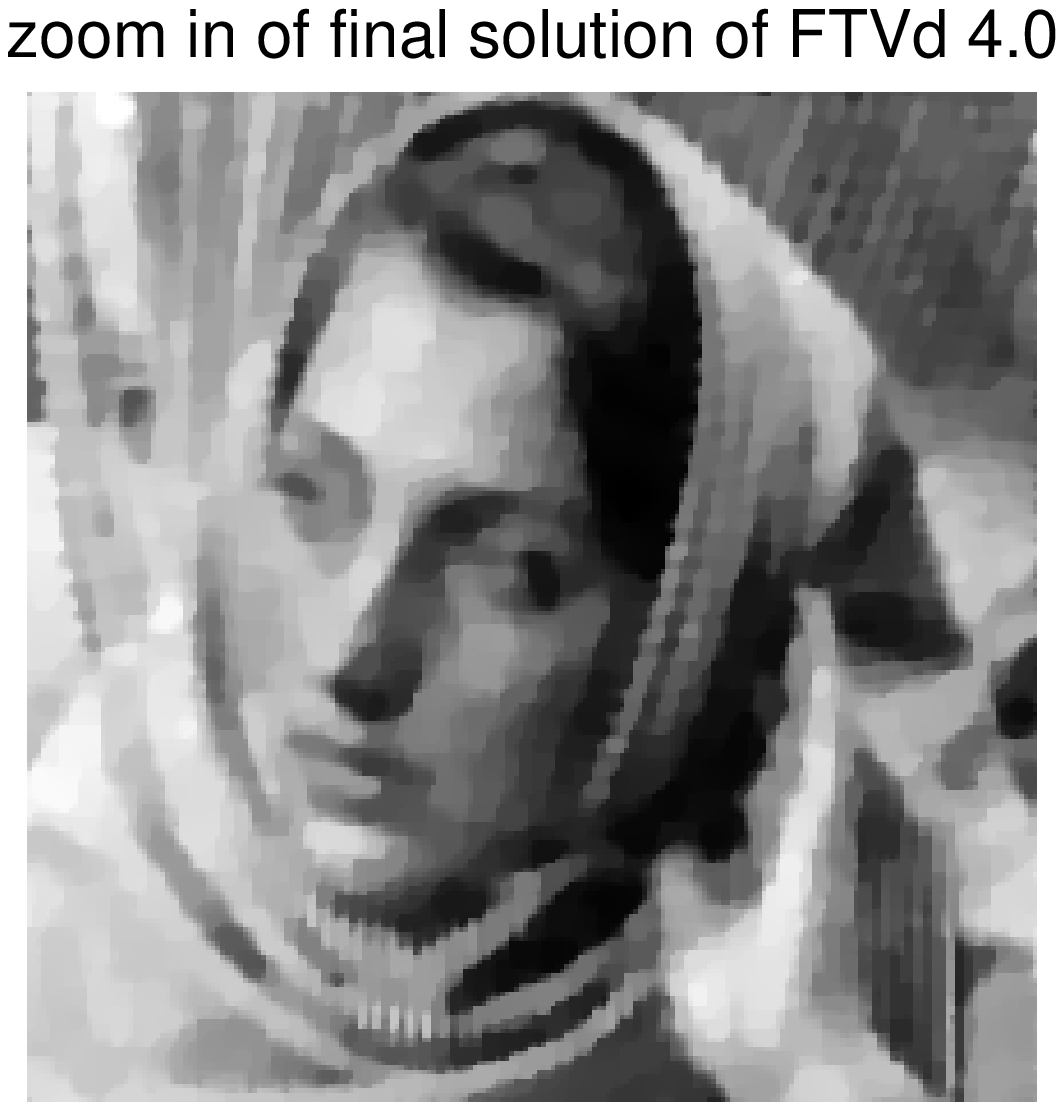}\\\vspace{-1cm}
    \caption{Image deblurring experiment of FTVd 4.0. Top left: original image. Top right: blurred noisy image. Second row: SNRs of the results of each iteration. Left of third row: intermediate result of highest SNR. Right of third row: the final solution. Four row: zoom in displays. We can see the intermediate result of the 3-th iteration has slightly high SNR and significant reduced staircase than the pure TV solution.}
    \label{fig:Barbara4}
\end{figure}

 Figure  \ref{fig:Barbara4}  is the result of FTVd 4.0 run on the image
Barbara. Compared with Figure \ref{fig:Barbara3}, the main difference is the second row, where
the SNRs of the intermediate results are corresponding to  every iteration (implicit continuation), instead of the explicit continuation step.  We can see the similar results as FTVd 3.0. The chosen intermediate result corresponding to the combined TV and Tikhonov regularizations \eqref{AppProb2} of highest SNR has significant reduced staircase artifacts than the final  solution, as showed in the third and fourth rows. 


\section{Conclusion}
In this paper, we have demonstrated that some of intermediate results of FTVd can empirically achieve better recovery quality than the final solution, which is expected to be the solution of the pure TV model \eqref{TV-reg}. We explain that FTVd in fact solves a series of combined TV and Tikhonov regularized  model \eqref{AppProb2}. Specifically,  either the continuation scheme in FTVd 3.0 or the method of multipliers in FTVd 4.0 can be reconsidered as a procedure to adjust the proportion of Tikhonov regularization compared with TV regularization from large to approaching null. Therefore, some intermediate results  empirically achieve better recovery quality than the final solution, in terms of reduced staircase effect, when a proper balance between these two reguluarizations has been achieved.


%



\section*{Acknowledgment}

This work was supported by the Natural Science Foundation of China, Grant
Nos. 11201054, 91330201 and by the Fundamental Research Funds for the Central Universities ZYGX2012J118.

\ifCLASSOPTIONcaptionsoff
  \newpage
\fi



\bibliographystyle{IEEEtran}
\bibliography{ADMRef,TV_ref,Bregman,ImageQualityEvaluator}
\end{document}